\documentclass[preprint]{elsarticle}

\usepackage{mypreamble}

\begin{document}

\begin{frontmatter}

%%% Title
\title{A hybrid machine learning framework for clad characteristics prediction in metal additive manufacturing}

\author[1,2]{Sina Tayebati\fnref{fn2}}
\author[1,2]{Kyu Taek Cho\fnref{fn1}}

%%% Affiliations
\address[1]{Department of Mechanical Engineering Northern Illinois University, DeKalb, Illinois, 60115 USA}
\address[2]{Advanced Research of Materials and Manufacturing Laboratory Northern Illinois University, DeKalb, Illinois, 60115 USA}

%%% First footnote
\fntext[fn1]{Corresponding author at: Department of Mechanical Engineering Northern Illinois University, DeKalb, Illinois, 60115 USA. \\Email address: \href{mailto:kcho@niu.edu}{kcho@niu.edu} (Kyu Taek Cho)}

%%% Second footnote
\fntext[fn2]{Present address: Department of Electrical and Computer Engineering University of Illinois at Chicago, Chicago, Illinois, 60607 USA.}

%%% Abstract
\begin{abstract}
During the past decade, metal additive manufacturing (MAM) has experienced significant developments and gained much attention due to its ability to fabricate complex parts, manufacture products with functionally graded materials, minimize waste, and enable low-cost customization. Despite these advantages, predicting the impact of processing parameters on the characteristics of an MAM printed clad is challenging due to the complex nature of MAM processes. Machine learning (ML) techniques can help connect the physics underlying the process and processing parameters to the clad characteristics. In this study, we introduce a hybrid approach which involves utilizing 
the data provided by a calibrated multi-physics computational fluid dynamic (CFD) model and experimental research for preparing the essential big dataset, and then uses a comprehensive framework consisting of various ML models to predict and understand clad characteristics. We first compile an extensive dataset by fusing experimental data into the data generated using the developed CFD model for this study. This dataset comprises critical clad characteristics, including geometrical features such as width, height, and depth, labels identifying clad quality, and processing parameters. Second, we use two sets of processing parameters for training the ML models: machine setting parameters and physics-aware parameters, along with versatile ML models and reliable evaluation metrics to create a comprehensive and scalable learning framework for predicting clad geometry and quality. This framework can serve as a basis for clad characteristics control and process optimization. The framework resolves many challenges of conventional modeling methods in MAM by solving t the issue of data scarcity using a hybrid approach and introducing an efficient, accurate, and scalable platform for clad characteristics prediction and optimization.
\end{abstract}

%%% Keywords
\begin{keyword}
\texttt Additive manufacturing \sep Machine learning  \sep Process map prediction\sep Clad geometry prediction \sep Hybrid modeling
\end{keyword}

\newcommand{\abbreviations}[1]{%
  \nonumnote{\textit{Abbreviations:\enspace}#1}}
\abbreviations{NN, Neural Network; KNN, K-Nearest Neighbor; GBR, Gradient Boosting Regression; GBC, Gradient Boosting Classification; RF, Random Forest; DT, Decision Tree; AB, AdaBoost; GPR, Gaussian Process Regression; SVR, Support Vector Regression, SVC, Support Vector Classification; Poly, Polynomial Regression; Lasso, Lasso Regression; Ridge, Ridge Regression.}

\end{frontmatter}

\begin{figure*}[ht!]
    \centering
    \includegraphics[width=\linewidth]{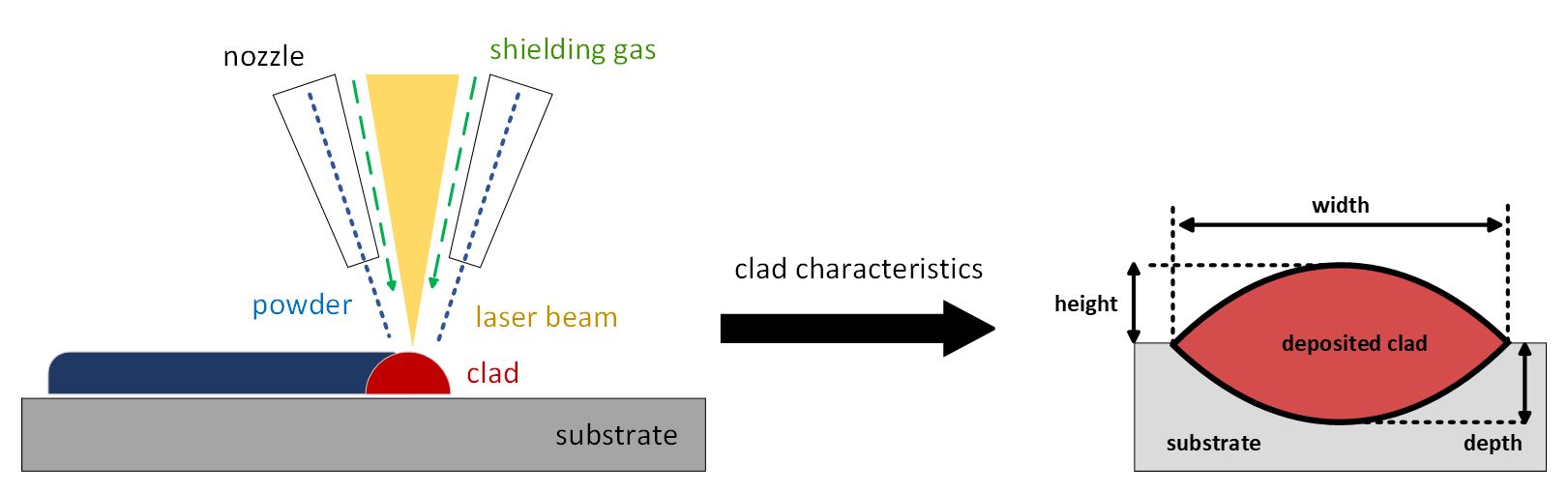}
    \caption{Schematic of the DED process and the clad characteristics.}
    \label{fig:am-process}
\end{figure*}

\section{Introduction}

Metal additive manufacturing (MAM) is a crucial and promising technology for producing high-performance parts used in the aerospace, automotive, and medical industries. Direct energy deposition (DED) is a specific MAM technique that enables free-form fabrication, remanufacturing, and modification of industrial parts. By depositing different materials layer by layer, DED enables the creation of heterogeneous metal parts in various sizes, with customizable properties \cite{gibson2015directed, ahn2021directed, tang2020review}. This technology involves using high-powered energy sources like electron beams or lasers to create a melt pool, where powder or wire feedstock is added to build the clad. Despite its commendable aspects, including flexibility, faster printing process, and capacity to produce large-scale parts, DED faces challenges in attaining optimal production speed, scale, and quality \cite{berman20123, wang2020machine, ian2015additive}. Ensuring high-quality parts requires regulating geometric characteristics during manufacturing, managing variables upstream and downstream of the process, and controlling factors like material properties, machine settings, and environmental conditions \cite{lee2017study}. Only by exercising proper control over these variables can consistent, high-quality parts be achieved using DED technology \cite{sciammarella2018processing, shim2016effect, li2017additive, izadi2020review}.

Analytical or numerical modeling is widely employed to investigate the characteristics of printed clad and establish optimal processing windows \cite{toyserkani2003three, rai2017simulation, bontha2009effects, gockel2014integrated}. For instance, Fathi et al. \cite{fathi2006prediction} developed a mathematical model to predict the melt pool depth and dilution of laser powder deposition, while Birnbaum et al. \cite{birnbaum2016simulating} utilized numerical modeling to predict the geometrical characteristics of specimens printed by powder-jet laser additive manufacturing. Wang et al. \cite{wang2020multi} reported a multi-physics model that employs the Finite Volume Method (FVM) to study the effects of physical phenomena underlying the DED process on the geometry of the printed clad. Additionally, successful implementations of empirical-statistical models have been reported to investigate the relationship between processing parameters and target property in the prints \cite{pant2019statistical,erfanmanesh2017empirical, jelvani2020empirical, nabhani2018empirical, zhang2019experimental}.

In recent years, due to the complexity of modeling MAM processes, a shift has occurred in the studies from relying solely on physics-based methods to adopting a hybrid approach that combines physics-based and data-driven models \cite{wang2020machine}. Machine learning (ML) models, trained using reliable data, offer a cost-effective alternative for advanced manufacturing \cite{marrey2019framework, karaboga2008performance, yildiz2013new, yildiz2009novel}. These ML algorithms, along with large datasets, enhance our understanding of AM intricacies and optimize processing parameters to achieve the desired quality in printed parts. Recent studies have demonstrated the potential of ML models in various AM applications. For example, Mehrpouya et al. \cite{mehrpouya2019prediction} developed a successful artificial neural network model to optimize laser parameters. Similarly, other studies have proposed ML algorithms to predict and control print defects \cite{lee2023explainable, akbari2022meltpoolnet, aoyagi2019simple, zhang2019hybrid, du2021physics} and improve geometrical properties of the melt pool, aiming to enhance the quality of the built specimens \cite{caiazzo2018laser, ren2021physics, biyikli2022single, liu2018geometry, xiong2014bead}.

\begin{figure*}[ht!]
    \centering
    \includegraphics[width=\linewidth]{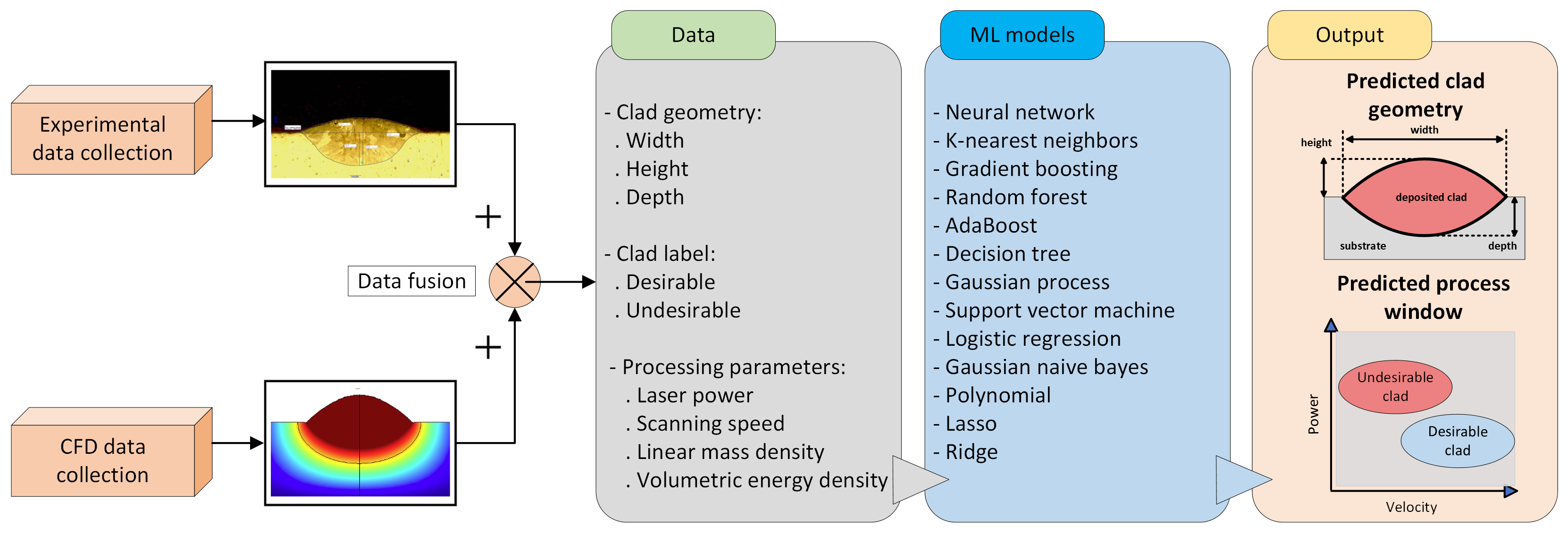}
    \caption{Hybrid data, features, ML models, and tasks implemented in our ML framework.}
    \label{fig:ml-framework}
\end{figure*}

However, the development of ML models faces various challenges, including the scarcity and variability of data due to expensive experiments and the existence of various AM machines. To address this, hybrid ML modeling approaches have been proposed, incorporating physics-based models to generate and augment the preliminary training data \cite{baturynska2018optimization, zhu2021machine, yin2011hybrid, yuguang2013improved}. For instance, Ren et al. utilized a finite element model to simulate the thermal field and build a large thermal history dataset for training their ML algorithm \cite{ren2020thermal}. Mohajernia et al. \cite{mohajernia2022machine} employed a multi-physics finite element model calibrated by experimental data to generate and provide a reliable dataset for ML training and testing. Another approach involves incorporating physics-based models not only to produce more data but also to enhance predictions by integrating physics knowledge within the model's loss function \cite{liu2021physics, kapusuzoglu2020physics}.

In order to bridge the existing knowledge gap and overcome the challenges in predicting various characteristics of MAM printed clad, this study introduces an innovative hybrid machine learning framework. The goal of this study is to develop a comprehensive solution that can be applied across different scenarios. To achieve this, the proposed workflow leverages an experimentally calibrated multiphysics computational fluid dynamics (CFD) model in conjunction with real-world experimental data, resulting in a rich and diverse training dataset for the ML models. The CFD model serves as a simulation tool that accurately captures the dynamics of the DED process, generating valuable information on the characteristics of the clad. This simulated data is then combined with experimental data, creating a large dataset that encompasses a wide range of scenarios and variations. By incorporating both simulated and real-world data, we ensure a robust and representative training set for the ML models. The ML models within this framework are specifically designed to optimize processing parameters and provide accurate predictions of the clad geometry. These models utilize advanced algorithms and techniques to analyze the vast amount of training data, identifying patterns and relationships that lead to improved predictions. By optimizing processing parameters, this platform enables the production of high-quality DED prints with enhanced precision and efficiency. The significance of this work lies in the comprehensive, scalable, and sustainable nature of this ecosystem. By addressing the challenge of data scarcity in machine learning, this study offers a solution that can be scaled to different manufacturing scenarios, providing accurate predictions for MAM processes. Moreover, this approach represents a substantial advancement over traditional physics modeling techniques, as it enhances generalization and accuracy by leveraging the power of machine learning.

    \section{Methodology}

In this study, Figure \ref{fig:ml-framework} presents the machine learning framework proposed, which summarizes various elements, including the dataset and its features, the hybrid method used for producing the data, the trained and tested ML models, and the output obtained from the framework.

To begin, this section discusses the process of collecting and preparing the dataset using the hybrid method. It also introduces the data features, which includes the machine setting and physics-aware features, as well as the data preprocessing methods and evaluation metrics applied in the study.

Additionally, the section highlights the ML models used in the study and their respective characteristics, as well as the process of hyperparameter optimization applied to these models. Overall, the framework provides a comprehensive approach to applying machine learning in additive manufacturing.

\begin{figure*}[ht!]
    \centering
    \includegraphics[width=\linewidth]{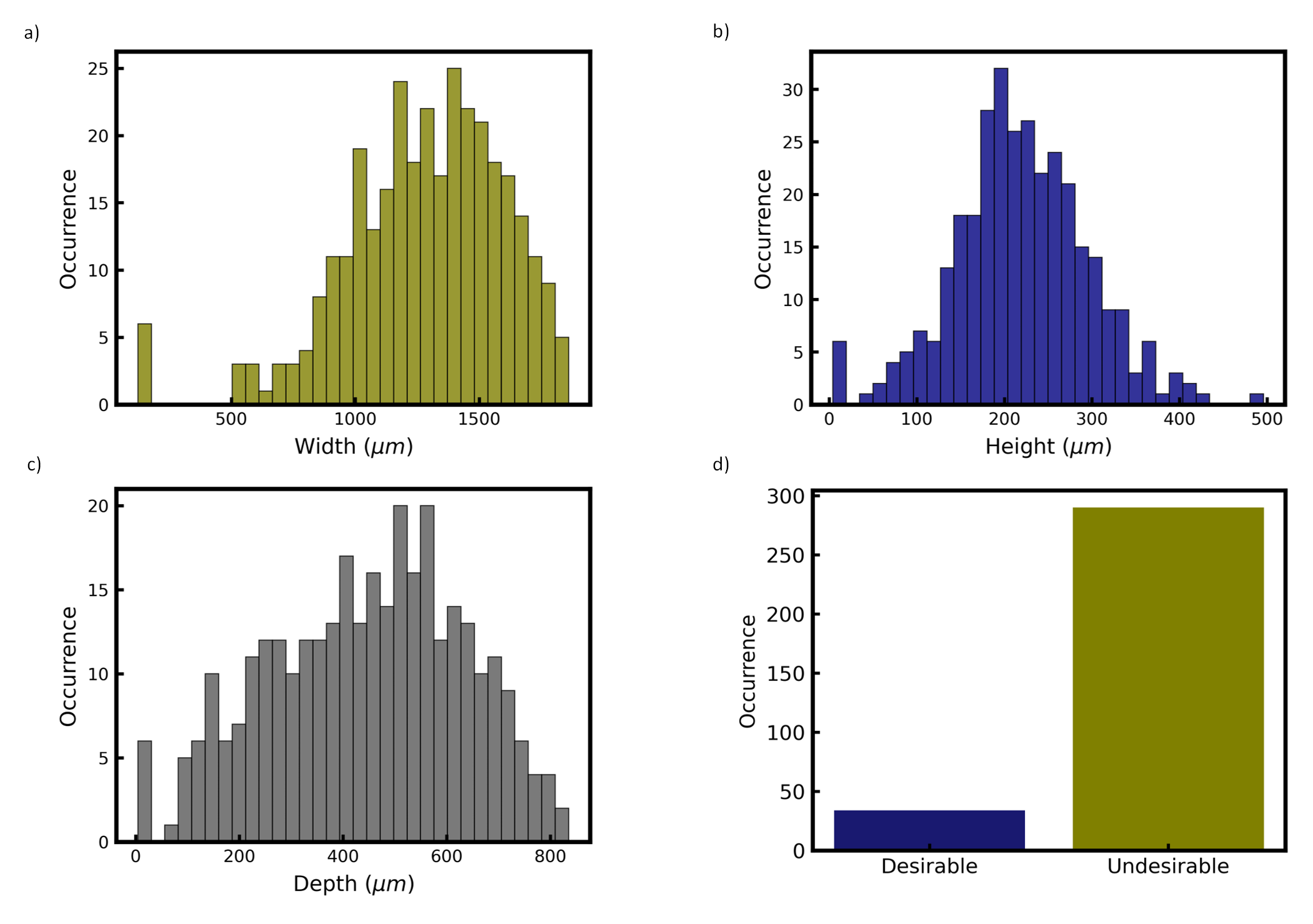}
    \caption{Distribution of clad features in our dataset, (a) "width" distribution, (b) "height" distribution, (c) "depth" distribution, (d) occurrence of clad quality labels.}
    \label{fig:occurrence}
\end{figure*}

\subsection{A hybrid machine learning architecture}

Machine learning models are powerful tools for predicting the performance of the AM process. However, their accuracy is highly dependent on the quality and quantity of data available to them. In contrast, metal additive manufacturing is a complex process, and gathering large amounts of experimental data across a wide range of processing parameters to serve the ML models, can be expensive and time-consuming. Therefore, here we devised a hybrid modeling approach to harness the potential of ML models for AM process modeling, while tackling the issue of data scarcity. We employed a calibrated computational fluid dynamics (CFD) model as a supplementary tool to generate vital data across a broad processing window. The integration of this CFD-generated data with experimental data formed the database for the machine learning framework. This approach enabled us to model the AM process at scale and make precise and dependable predictions. This dataset includes 325 data points comprised of 90 single clad data collected from experimental study \cite{cho2023investigation}, and 235 data points generated by CFD model within a wide processing window. In this dataset, each data point consists of information such as the processing parameters and the associated clad features i.e., width, depth, height, and quality label. The distributions of the targeted clad features in our dataset are shown in \autoref{fig:occurrence}.

\subsection{Data from numerical model and experimental setup}

The direct energy deposition simulation was conducted using a 2D symmetric computational fluid dynamics model developed with COMSOL Multiphysics 5.3 (COMSOL, Burlington, MA, USA). This model was specifically designed to simulate the generation of the melt pool and the resulting geometrical features of the clad. To achieve this, the model solved a system of differential equations, including conservation of energy, mass, and momentum equations. The physics of the model accounted for natural convection and the Marangoni effect. The simulation incorporated temperature-dependent thermal properties of stainless steel 316 L, such as conductivity, viscosity, and specific heat. Additionally, a modified dynamics viscosity approach was employed to model the phase change interface. The generation of the clad was simulated using a deformed geometry function through a moving mesh based on an arbitrary Lagrangian–Eulerian method (ALE) in COMSOL, which allows simulating mass addition into the melt pool and predicting the clad growth and generation. The laser heat source was represented by a 2D Gaussian distribution intensity, and an interaction time was calculated to account for the effect of a moving heat source. This interaction time depended on the velocity and radius of the laser beam. By incorporating the interaction time, the simulation eliminated the influence of melt pool lag and enabled the representation of the laser cladding process in a 2D cross-sectional domain. The CFD model was discretized using quadrilateral elements, with element sizes controlled at 13 \textmu for the melt pool generation domain and 100 \textmu near the fixed temperature boundaries. Mesh independence was studied by checking the effect of mesh size on the simulation results as the mesh size decreased, and the element sizes at which the computation results were not changed anymore were utilized for this modeling study. Additionally an experimental study was conducted to validate the numerical model. A comprehensive description of the numerical modeling and experimental methodology can be found in our previous publication \cite{cho2023investigation}.

\subsection{Feature engineering}

The process of feature engineering involved creating and defining suitable features in the dataset to be used as inputs for ML models in regression and classification tasks. Given the complexity of the AM process physics, selecting the right features with statistical significance is crucial for achieving the highest accuracy in predictions. In this study, we selected a set of machine setting features that can be directly controlled by the equipment operator. These features are commonly used in the literature as they can be easily defined and utilized through experimental work. Additionally, we identified physics-aware features, which are parameters that capture more detailed information about the underlying physics of the AM process. By incorporating both the machine setting features and physics-aware features as inputs to the ML models, we aimed to achieve higher accuracy and more robust predictions \cite{du2021physics, ren2021physics}. The following sections provide more detailed information about these features.

\subsubsection{Machine setting features}

In metal additive manufacturing, melt pool properties are highly correlated with the features defined and controlled directly through the machine and within the experiment process. The constructed dataset consists of two process parameters including the laser power $(W)$,  and laser scanning velocity $(mm/s)$  as the baseline machine setting features that were defined and controlled within the experiments and CFD modeling.

\subsubsection{Physics aware features}

Previous research in the literature suggests that utilizing only machine setting features is often insufficient in uncovering the relationship between the underlying physics of the AM process and the target variables required for accurate prediction using machine learning (ML) models. The lack of a clear linkage between these features can result in low accuracy, bias, and misleading predictions \cite{du2021physics, kapusuzoglu2020physics}. Additionally, machine setting features are specific to the AM machine used in the study, which may not be transferrable to other studies with different machine technologies, resulting in limitations in their applicability across different studies \cite{liu2021physics}. Therefore, novel approach involves defining machine-independent "physics-aware" features as additional input variables for ML models. These features, which could be derived from experimental and computational data, integrate with machine setting features, addressing transferability limitations and enhancing understanding of AM physics. This approach, known as "physics-informed" ML modeling, improves predictions of melt pool and clad properties in additive manufacturing. \cite{ren2021physics}. In our study, the volumetric energy density $(J/mm^3)$ and linear mass density $(g/mm)$ were selected as the physics-aware features.

\begin{table*}
\centering
\caption{Featurization employed in the ML framework}
\label{tab:features}
\begin{tabular}{lcc}
\toprule
Machine features  &  Physics aware features \\
\midrule
Laser power $(W)$   &   Volumetric energy density $(J/mm^3)$ \\
Laser scanning speed $(mm/s)$  &   Linear mass density $(g/mm)$ \\
\bottomrule
\end{tabular}
\end{table*}

Volumetric energy density plays a crucial role in understanding and controlling potential defects in cladding process. For instance, low laser power and fast laser scanning speed results in too little energy input per unit length which provides insufficient energy to form a proper molten pool for cladding process. This phenomena often results in a discontinues molten pool that generates several balls that are considered a defect. The volumetric energy density ($E$) is expressed as \cite{debroy2018additive, wei2021mechanistic}.

\begin{equation}\label{eq:14}
E = \frac{P}{v (\pi r^2)}
\end{equation}
Where $P$ is laser power, $v$ is laser scanning speed, and $r$ is laser beam radius. This represents the amount of energy provided from the laser heat source per unit volume of the deposited material.

Linear mass density is also an essential feature for producing repeatable structures with minimum potential defects such as porosity. For instance, at high linear mass density with high rate of powder feed rate, the contact angle of the deposited clad increases, which results in an undesirable clad geometry causing gas entrapment between the clad and eventually increasing the porosity level \cite{ferguson2015semi}. The linear mass density ($\lambda_{m}$, ratio of powder feed rate to laser travel speed) is expressed as.

\begin{equation}\label{eq:15}
\lambda_{m} = \frac{\dot{m}}{v_{l}}
\end{equation}
Where $\dot{m}$ is the powder mass flow rate, and $v_{l}$ is the laser travel speed. This represents the maximum amount of powder mass deposited per unit length of laser travel. The summary of the parameters utilized for developing the ML framework are listed in \autoref{tab:metrics}.

\subsection{Data preprocessing and metrics}

Since the input features in our dataset are not in the same range and are highly heterogenous, input features of our dataset were normalized using the min-max normalization method expressed as.

\begin{equation}\label{eq:16}
   x_{normalized} = \frac{x-min(x)}{max(x) - min(x)}
\end{equation}

For each task in this study, different metrics were used for evaluating the performance of different models and a splitting pattern was utilized for training and testing. \autoref{tab:metrics} summarizes this information for each dataset and associated task.

Verification of the performance of machine learning models in unseen data is a critical aspect of ML modeling studies. To verify ML model performance on unseen data, we split the dataset into training and testing subsets using an 80:20 ratio \cite{caiazzo2018laser}. Additionally, we employed k-fold cross-validation, randomly shuffling the dataset and evaluating the models' average accuracy over $k$ iterations \cite{anguita2012k}. Furthermore, different metrics were utilized for evaluating and comparing the performance of the ML models for different tasks. For the regression task (predicting the geometry of the melt pool and clad) the list of utilized metrics includes mean absolute error (MAE) and $R^2$ coefficient of determination. For the classification task (predicting the process map), the evaluation metrics are accuracy, and the area under the receiver operating characteristics curve (AUC-ROC).

\begin{table*}[ht!]
\centering
\caption{Our framework dataset details including tasks, splitting methods, and metrics}
\label{tab:metrics}
\resizebox{12cm}{!} 
{
\begin{tabular}{lccc}
\toprule
Category  &  Tasks  &  Split  &  Metric \\
\midrule
Width of the clad   &   Regression  &  Random  & $R^2$ - MAE \\
Height of the clad  &   Regression  &  Random  & $R^2$ - MAE  \\
Depth of the clad   &   Regression  &  Random  & $R^2$ - MAE   \\
Clad Classification   &   Classification  &  Random  & Classification accuracy - AUC-ROC \\
\bottomrule
\end{tabular}
}
\end{table*}

\subsection{Models}

In this research study, we have developed and tested several different machine learning models on the previously explained dataset. These models were developed for different tasks and best serving models were chosen to be used based on the performance comparison. Each of these employed machine learning models will be presented and briefly discussed below.

\subsubsection{Gaussian Naïve bayes ‘GNB’}

Gaussian naïve bayes is a very simple yet competitive probabilistic machine learning model used for classification. The core of this algorithm is built on the Bayes theorem, and it relies on the naïve assumption of conditional independency of all features given their classes \cite{berrar2018bayes}. Naïve bayes classifiers are extremely fast learning models due to the simplicity of the algorithm. In this study, naïve bayes is utilized for our classification task.

\subsubsection{Polynomial linear regression ‘Poly’}

Polynomial regression is a widely used model with assumptions very similar to the linear regression models. In other words, the polynomial regression model assumes that there are more than one independent variables $X_{j}$ correlating with the variable $Y$, the errors are normally distributed for each observation, and that variables  $X_{j}$ are fixed and the  only randomness in $Y$ comes from the error term \cite{ostertagova2012modelling}. In this study, the polynomial regression was used for the regression task.

\subsubsection{Lasso linear regression ‘Lasso’}

Lasso regression is a variant of linear regression focused towards solving the multicollinearity issue in the data, or case where we need to automate certain parts of the model selection such as variable elimination. Lasso regression performs the ‘L1’ regularization which adds penalties equal to the absolute value of the magnitude of the coefficient which can result in sparse models or may cause elimination of some of the coefficients which eventually creates models with less complexity \cite{ranstam2018lasso}. In this research, Lasso regression was only utilized for the regression task.

\subsubsection{Ridge linear regression ‘Ridge’}

Ridge regression is a regularized linear regression model that addresses the problem of overfitting in high-dimensional datasets by adding a penalty term to the cost function. The penalty term, which is proportional to the square of the magnitude of the coefficients, shrinks the coefficients towards zero, reducing their impact on the prediction \cite{arashi2019theory}. This results in a more stable and reliable model. Ridge regression is particularly useful in situations where the number of predictors is larger than the number of observations or when there is multicollinearity among the predictors. In this study, the ridge regression was utilized only for the regression task.

\subsubsection{Support vector machine (‘SVR’ and ‘SVC’)}

Support vector machines (SVM) are versatile machine learning algorithms used for classification and regression tasks. Support vector regression (SVR) minimizes deviation from the insensitivity parameter, while support vector classification (SVC) finds the optimal hyperplane for class separation, maximizing the margin between classes. SVMs handle high-dimensional data and can handle linear and non-linear separable datasets using various kernel functions. These models can be regularized to prevent overfitting and have proven effective in diverse applications \cite{gholami2017support}.

\subsubsection{Gaussian process model (‘GPR’ and ‘GPC’)}

Gaussian processes regression (GPR) and Gaussian process classification (GPC) are non-parametric machine learning algorithms used for regression and classification tasks. These models are a generalization of the Gaussian probability distribution over possible functions \cite{bousquet2011advanced}. This model’s output is a probability distribution over all possible output values due to its Bayesian approach. These models show great performance on small datasets. However, with increasing amount of datapoints, Gaussian process can be computationally expensive. In this study, Gaussian process model was utilized for regression and classification.

\subsubsection{Decision tree ‘DT’}

The decision tree is a widely used supervised machine learning model for both regression and classification tasks. This non-parametric algorithm is particularly well-suited for handling large datasets efficiently. In this approach, observations are segmented into branches of a tree, consisting of decision nodes and leaves, or used to predict a response vector $Y$ associated with an input matrix $X$ \cite{song2015decision}. The leaves contain the outcome or decision of the model for labeling an observation by a certain class. The decision tree was utilized for regression and classification tasks in this research study.

\subsubsection{Random forest ‘RF’}

Random forest models have proved great capabilities for classification and regression specially for large scale datasets. Random forest follows the ensemble learning and utilizes a group of randomized decision trees to be trained on random batches of dataset. This model presents the final prediction as the average of the outputs and the majority vote of the decision trees for regression and classification tasks respectively \cite{biau2016random}. In this study, random forest was used for both the regression and classification tasks.

\subsubsection{Gradient boosting model (‘GBR’ and ‘GBC’)}

Gradient boosting is an effective ensemble learning technique that combines weak decision tree models to create a strong predictive model. By iteratively adding new models to the ensemble, gradient boosting corrects errors and minimizes the loss function using gradient descent \cite{natekin2013gradient}. It is widely used for regression and classification tasks due to its scalability, high accuracy, and capability to handle linear and non-linear relationships.

\begin{table*}[ht!]
\centering
\caption{Hyperparameters and the range studied for the ML model in clad depth prediction.}
\label{tab:hyperparameter-reg}
%\small
\resizebox{12cm}{!} 
{
\begin{tabular}{cccc}
\toprule
  &    &  Depth regression  &    \\
\midrule
Models  &  Hyperparameter  &  Range  &   Value \\
\midrule
     NN    & number of hidden layers  & 1-6   &  2 \\
           & activation function  & [idendity, logistic, tanh, relu]   &  relu \\
           & number of neurons &  [256, 128, 64, 32]  &  (128, 64) \\
           & optimizer  & [lbfgs, SGD, adam]   &  lbfgs \\
           & regularization  &  [L1, L2]  &  L2 \\
           & learning rate  &  [$10^{-2}$, $10^{-3}$, $10^{-4}$]  &  $10^{-3}$ \\
\hline
     KNN   &  number of neighbors  &  10 – 1000  & 10  \\
           &  leaf size  &  10 – 1000  & 560  \\
           &  algorithm & [auto, ball-tree, kd-tree, brute]   &  brute \\
           &  weights  &  [distance, uniform]  & distance  \\
           &  metric  & [Minkowski, Euclidean, Manhattan]   &  manhattan  \\
\hline
     GBR   &  number of estimators  &  50 – 2000  &  916 \\
           &  learning rate  & $10^{-2}$ – 1   &  $10^{-2}$ \\
           &  max features  & [auto, sqrt]   &  sqrt \\
           &  max depth  &  10 – 110  &  100 \\
           &  min sample split  &  [2, 5, 10]  & 2  \\
           &  min sample leaf  &  [1, 2, 4]  &  1 \\
\hline
     RF    &  number of estimators  &  50 – 2000  & 1783  \\
           &  max features  &  [auto, sqrt]  & sqrt  \\
           &  max depth  &  10 – 110  &  10 \\
           &  min sample split  &  [2, 5, 10]  &  5 \\
           &  min leaf split  &  [1, 2, 4]  &  1 \\
\hline
     DT    &  max depth  &  10 – 110  &  60 \\
           &  max features  &  [auto, sqrt, log2]  &  sqrt \\
           &  min sample split  &  [2, 5, 10]  &  2 \\
           &  min leaf split  &  [1, 2, 4]  & 1  \\
\hline
     AB    &  number of estimators  &  50 – 2000  &  50 \\
           &  learning rate  &  $10^{-2}$ – 1  &  1 \\
           &  loss  &  [linear, square, exponential]  &  square \\
\hline
     GPR   &  kernel  &  [RBF (), WhiteKernel (), Matern ()  & 1**2 * RBF (length-scale=1)   \\
           &    &  , ConstantKernel (), ExpSineSquared()]  & + WhiteKernel(noise-level=1)  \\
           &  alpha  &  [$10^{-3}$, $10^{-2}$]  &  $10^{-2}$ \\
\hline
     SVR   &  kernel  &  [linear, poly, rbf]  & rbf  \\
        &  C  &  1 – 1000  &  290 \\
\hline
     Poly  &  degree  &  2 - 5  &  4 \\
\hline
     Lasso &  alpha  &  $10^{-4}$ - 1  &  0.0045 \\
\hline
     Ridge &  alpha  &  $10^{-4}$ - 1  &  0.012 \\
\bottomrule
\end{tabular}
}
\end{table*}

\begin{table*}[ht!]
\centering
\caption{Hyperparameters and the range studied for the ML model in classification.}
\label{tab:hyperparameter-clf}
\resizebox{12cm}{!} 
{
\begin{tabular}{cccc}
\toprule
  &    &  Clad classification  &    \\
\midrule
Models  &  Hyperparameter  &  Range  &   Value \\
\midrule
     NN   & number of hidden layers  & 1-6   &  4 \\
           & number of neurons &  [256, 128, 64, 32]  &  (256, 128, 64, 32) \\
           & optimizer  & [lbfgs, SGD, adam]   &  SGD \\
           & activation  &  [identity, logistic, tanh, relu]  &  idendity \\
           & alpha  &  [$10^{-4}$, $10^{-3}$, $10^{-2}$]  &  $10^{-3}$ \\
           & learning rate  &  [constant, invscaling, adaptive]  &  constant \\
\hline
     LR    &  penalty  &  [L1, L2]  &  L2 \\
           &  C  &  $10^{-5}$ – 10  &  4 \\
           &  solver &  [lbfgs, liblinear, newton-cg, newton-cholesky, sag, saga]  &   bfgs \\
\hline
     KNN   &  number of neighbors  &  1 – 20  &  15 \\
           &  weights  &  [uniform, distance]  &  distance \\
           &  algorithm & [auto, ball-tree, kd-tree, brute]   &  brute \\
           &  metric  & [minkowski, Euclidean, Manhattan]   &  minkowski  \\
\hline
     GBC   &  number of estimators  &  50 – 1000  &  50 \\
           &  learning rate  & $10^{-2}$ – 1   &  0.5 \\
           &  max features  & [auto, sqrt, log2]   &  log2 \\
           &  max depth  &  2 – 10  &  2 \\
           &  subsample  &  0.1 – 0.9  & 0.48  \\
\hline
     RF    &  number of estimators  &  50 – 1000  & 894  \\
           &  max features  &  [auto, sqrt, log2]  & sqrt  \\
           &  max depth  &  10 – 110  &  10 \\
           &  min sample split  &  [2, 5, 10]  &  2 \\
           &  min leaf split  &  [1, 2, 4]  &  4 \\
           &  criterion  &  [gini, entropy, log-loss]  &  entropy \\
\hline
     DT    &  max depth  &  10 – 1000  &  890 \\
           &  max features  &  [auto, sqrt, log2]  &  log2 \\
           &  criterion  &  [gini, entropy]  &  gini \\
           &  min sample split  &  [2, 5, 10]  &  2 \\
           &  min leaf split  &  [1, 2, 4]  & 4  \\
\hline
     AB    &  number of estimators  &  50 – 1000  &  577 \\
           &  learning rate  &  $10^{-2}$ - 1  &  0.02 \\
\hline
     GPC   &  kernel  &  [RBF (), WhiteKernel(), Matern ()  &  1**2 * Matern  \\
           &    &  , ConstantKernel (), ExpSineSquared()]  &  (length-scale=1, nu=0.5) \\
           &  optimizer  &  [L-BFGS-B, L-BFGS, CG, NCG]  &  L-BFGS-B \\
\hline
     SVC   &  kernel  &  [linear, poly, rbf, sigmoid]  & rbf  \\
           &  C  &  10 – 1000  &  470 \\
           &  gamma  &  [scale, auto, 0.1, 1.0, 10.0]  &  1 \\
           &  degree  &  2 - 5  &  3 \\
\hline
     GNB   &  var-smoothing  &  $10^{-10}$ – $10^{-3}$  & $4\times10^{-4}$  \\

\bottomrule
\end{tabular}
}
\end{table*}

\subsubsection{AdaBoost ‘AB’}

AdaBoost (Adaptive Boosting) is a widely used boosting algorithm for classification and regression tasks in supervised machine learning. It combines multiple weak learners to create a strong classifier or regressor. The algorithm iteratively assigns higher weights to misclassified samples, trains new weak learners on reweighted data, and continues until a predetermined number of iterations or convergence. The final model is a weighted sum of the weak learners based on their performance during training \cite{an2010new}.

\subsubsection{K-nearest neighbor ‘KNN’}

The k-nearest neighbor is a famous algorithm utilized for regression and classification tasks. K-nearest neighbors are supervised non-parametric models meaning that it does not assume any specific probability distribution for the input data. These models work on small datasets but can be computationally expensive for large datasets. K-nearest neighbor algorithm aims to find the nearest pattern to a target pattern to classify unlabeled datapoints by assigning them to the most similar labeled observation in the target pattern \cite{zhang2016introduction}, or tries to map patterns to continuous labels for regression purposes \cite{kramer2013dimensionality}.

\subsubsection{Logistic regression ‘LR’}

Logistic regression is a statistical method commonly used for binary classification. It assumes linearity of independent variables, independence of observations, and no multicollinearity. It utilizes a logistic function (Sigmoid function) to model a binary dependent variable in a generalized linear model \cite{sperandei2014understanding}. In this study, logistic regression was employed for classification purposes.

\subsubsection{Neural network ‘NN’}

Neural networks are widely used in machine learning for regression and classification tasks. They excel at approximating complex nonlinear functions due to their universal function approximator property \cite{borisov2022deep}. Mimicking the interconnected neurons in the human brain, neural networks consist of input, hidden, and output layers with fully connected neurons. In this study, neural network models were employed for regression and classification.

\subsection{Hyperparameter optimization}

Optimizing hyperparameters is crucial for accurate and efficient machine learning models. It involves selecting the best hyperparameters that yield optimal performance. The choice of hyperparameters significantly impacts the model's learned parameters and final performance. In this study, we employed the Randomized Search method \cite{bergstra2012random} to identify the optimal hyperparameters for our regression and classification models. This method randomly samples the hyperparameter space, trains the model for each set of sampled hyperparameters, and evaluates performance using a validation set. The Randomized Search method aims to optimize an objective function, such as validation mean squared error for regression or validation accuracy score for classification. Compared to other optimization techniques, Randomized Search is efficient and less likely to get stuck in sub-optimal regions. In \autoref{tab:hyperparameter-reg}, the hyperparameters explored for the machine learning models predicting the clad depth are presented in detail. The optimized ML hyperparameters for clad width and height are reported in supplementary material. The details of the hyperparameters studied for the machine learning models classification task are presented in \autoref{tab:hyperparameter-clf}.

For the neural network model, six hyperparameters are optimized: number of hidden layers and nodes per layer, activation function, optimizer, learning rate, and regularization parameter alpha \cite{goodfellow2016deep, rumelhart1986learning}. Different combinations of hidden layers with neuron numbers of 256, 128, 64, and 32 are explored. An optimal value for these parameters is crucial to balance model complexity and overfitting. Four activation functions ('identity', 'logistic', 'tanh', and 'relu') are investigated. Three optimizers (lbfgs, SGD, and Adam \cite{kingma2014adam}) are explored to minimize the loss function. Alpha, a regularization parameter, ranges from $10^{-2}$ to $10^{-4}$ to encourage a less complex and generalizable model. Learning rate, ranging from $10^{-2}$ to $10^{-4}$, determines the step size for weight and bias updates.
For logistic regression, the regularization parameter C (inverse of regularization strength) is optimized in the range of $10^{-5}$ to 10. Both L1 and L2 penalty functions are explored. Various solvers ('lbfgs', 'liblinear', 'newton-cg', 'newton-cholesky', 'sag', and 'saga') are investigated to optimize the cost function \cite{li2014efficient, schmidt2017minimizing, simon2013blockwise, defazio2014saga}.

For the k-nearest neighbor model, the number of neighbors and leaf size are crucial hyperparameters. The range of 10 to 1000 is explored for the number of neighbors, where a smaller value provides a more flexible model. Leaf size determines the subdivision stopping point during tree construction. Different algorithms ('ball-tree', 'kd-tree', and 'brute') are explored to optimize the loss function \cite{goldberger2004neighbourhood, brearley2022knn}. Uniform and distance-based weight schemes are investigated, and three metrics (Minkowski, Euclidean, Manhattan) are explored to compute the distance between neighbors.

For the random forest and gradient boosting models, the hyperparameter 'number of estimators' determines the number of decision trees used. A range of 50 to 2000 trees is explored to find the optimal value for these models \cite{cutler2012random}. In addition, for the decision tree model \cite{breiman2017classification}, random forest, and gradient boosting, 'max depth' defines the maximum depth of each decision tree, and 'max features' limits the number of features considered for splitting at each tree node. 'Max depth' is explored in the range of 10 to 110, and 'max features' can be 'auto', 'sqrt', or 'log2'. The 'Min sample split' determines the minimum number of samples required to split at an internal node, explored in the range of 2 to 10, while 'min sample leaf' is the minimum number of samples required at a leaf node, searched in the range of 1 to 4. AdaBoost, another ensemble model, relies on the number of estimators, which determines the number of weak learners combined. A range of 50 to 2000 is explored. Three loss functions ('linear', 'square', 'exponential') are investigated using a learning rate from $10^{-2}$ to 1 to minimize the loss \cite{hastie2009multi}.

Support vector machine (SVM) models require optimization of the kernel and regularization parameter C. Four kernel functions were tested: linear, polynomial, sigmoid, and radial basis function (rbf). The linear kernel works for linearly separable datasets, while non-linear kernels (polynomial, rbf, sigmoid) transform the data into a higher dimension to achieve separability. The regularization parameter C controls the penalty for misclassified points. Smaller C values allow more misclassifications, while larger values restrict the decision boundary and may increase the risk of overfitting. The C range of 1 to 1000 was explored for optimization \cite{smola2004tutorial}.

\begin{figure*}[ht!]
    \centering
    \includegraphics[width=\linewidth]{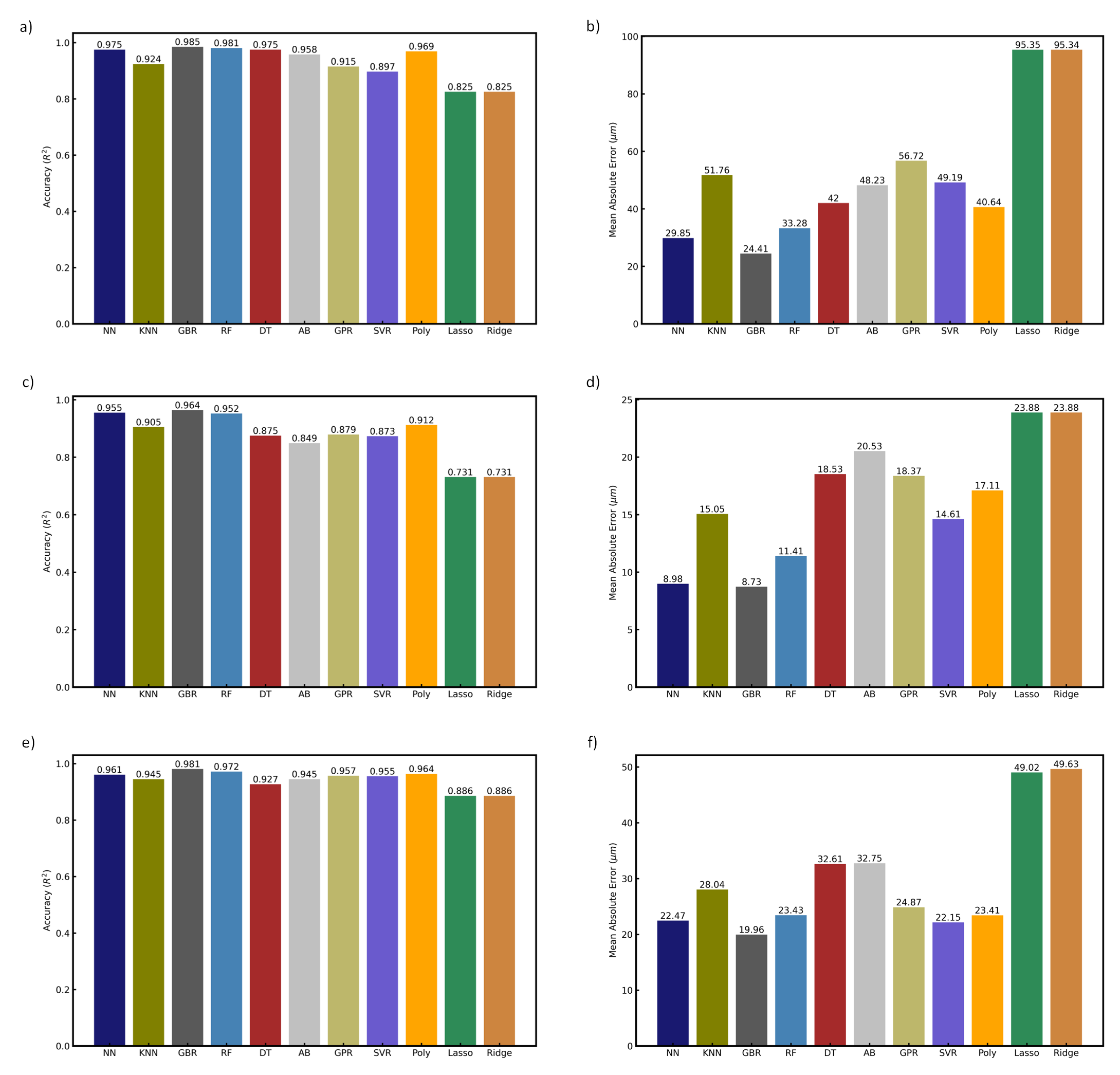}
    \caption{Benchmark performance comparison for prediction of geometrical characteristics of the single clad: (a, b) width prediction accuracy and MAE results, (c, d) comparison between the height prediction accuracy and MAE results, (e, f) comparison between the depth prediction accuracy and MAE results.}
    \label{fig:r2-mae}
\end{figure*}

The Gaussian process model is a kernel-based machine learning algorithm that requires the optimization of both the kernel function and the regularization parameter. The kernel determines the prior distribution over the functions and controls the complexity and degree of correlation between variables. Different kernels can capture various patterns in structured data, such as linearity, non-linearity, or periodicity. To find the optimized function, different combinations of kernels, including 'rbf', 'Matern', 'WhiteKernel', 'ExpSineSquared', and 'ConstantKernel', were tested \cite{wilson2013gaussian}. Additionally, the regularization parameter alpha was studied for the gaussian process regression model in a range if $10^{-3}$ to $10^{-2}$. Regularization parameter controls the level of noise present in the observed data and determines the smoothness of the model's prediction, meaning that smaller values of alpha result in a model that fits the data more closely. Moreover, three optimizers i.e., Limited-memory Broyden–Fletcher–Goldfarb–Shanno (L-BFGS), Limited-memory Broyden Fletcher Goldfarb–Shanno with Bound constraints (L-BFGS-B), Conjugate Gradient (CG), and Newton Conjugate Gradient (NCG), were studied for the gaussian process classifier model \cite{rasmussen2006gaussian, schulz2018tutorial}.

The Gaussian Naïve Bayes model involved exploring a range of $10^{-10}$ to $10^{-3}$ for the 'var-smoothing' hyperparameter. This hyperparameter controls the amount of smoothing applied to feature variances to avoid division by zero errors. For linear lasso and ridge regressions, the regularization parameter alpha was investigated in the range of $10^{-4}$ to 1 to prevent overfitting \cite{friedman2010regularization, kim2007interior}. In polynomial regression, the degree of the model is a hyperparameter chosen based on problem complexity. The polynomial regression model was optimized by exploring degree values in the range of 2 to 4 \cite{cheng2018polynomial, ostertagova2012modelling}.

    \begin{figure*}[ht!]
    \centering
    \includegraphics[width=\linewidth]{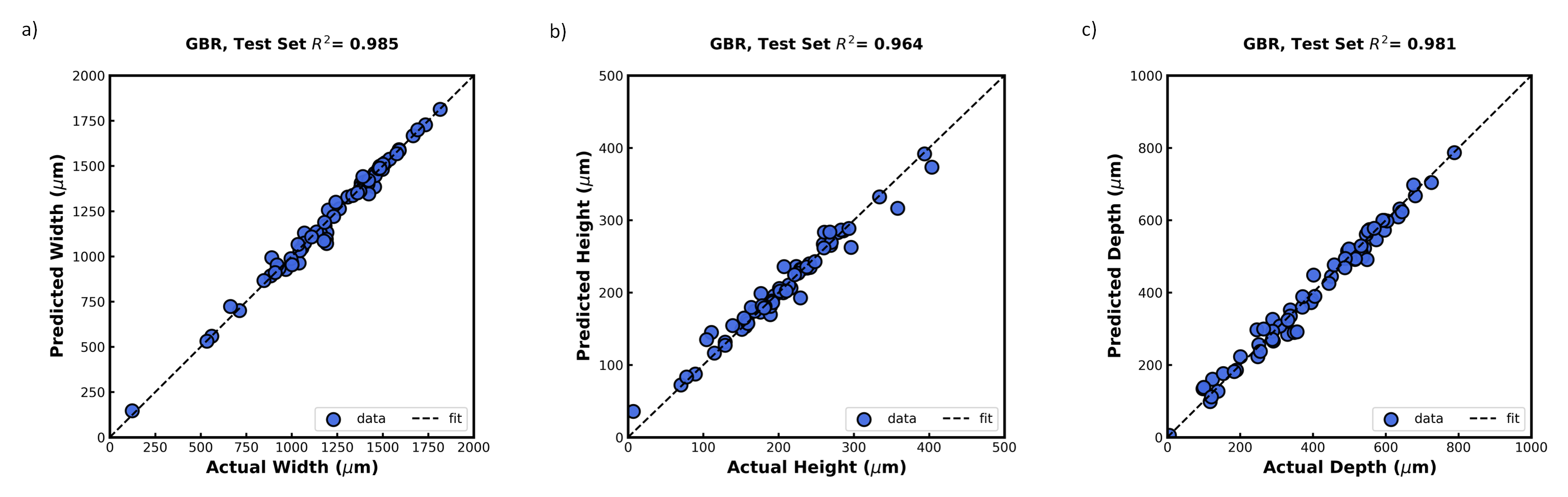}
    \caption{Predicted clad geometry plotted against the actual grand truth geometry values: (a) width prediction using ‘Gradient Boosting’, (b) height prediction using ‘Gradient Boosting’, (c) depth prediction using ‘Gradient Boosting’.}
    \label{fig:r2}
\end{figure*}

\section{Results and discussion}

The purpose of this section is to present the outcomes of the benchmarked models within the proposed framework for the two primary tasks: predicting the geometry of single clads and determining the optimal process map. To assess the performance of the models for each task, multiple metrics were utilized, and the most effective models for each were determined. Further elaboration will be provided in the succeeding sections.

\subsection{Clad geometry prediction}

Eleven machine learning (ML) models were designed and implemented to predict the clad geometrical features, including depth, width, and height. The employed ML models, namely 'Neural Network', 'Gradient Boosting', 'Random Forest', 'Decision Tree', 'AdaBoost', 'Support Vector Machine', 'Gaussian Process', 'K-nearest Neighbors', 'Polynomial Regression', 'Ridge Regression', and 'Lasso Regression', were evaluated for their performance in terms of accuracy and error, measured by the coefficient of determinant $(R^2)$ and mean absolute error (MAE $\mu m$), respectively. The input features for the models were selected from four categories, namely machine features and physics aware features, as listed in \autoref{tab:features}. Hyperparameter optimization was conducted on each ML model to ensure optimal performance. The detailed results of the performance evaluation of the models are presented in the subsequent sections.

The accuracy $(R^2)$ and MAE results of the ML models were assessed on the clad width prediction task and have been reported (Fig. \ref{fig:r2-mae} a-b). We ran the optimized ML models, and the best result was achieved with the gradient boosting model which yielded 0.985 accuracy and an MAE of 24.41 \textmu m for a combination of four machine setting and physics aware features mentioned in previous sections. The second and the third place in the list of best performing models for predicting the width belongs to random forest and neural network models respectively. Random forest yielded an accuracy of 0.981 and an MAE of 33. 28 \textmu m, and the neural network model demonstrated an accuracy of 0.975\% and an MAE of 29.87 \textmu m for predicting the clad width.

The accuracy ($R^2$) and MAE results of the ML models in the framework for predicting the clad height has been illustrated (Fig. \ref{fig:r2-mae} c-d). The best accuracy was achieved using the four input features for gradient boosting model with 0.964 accuracy and an MAE of 8.73 \textmu m. The second place in the list of best performing models for predicting the height belongs to the neural network with an accuracy of 0.955 and an MAE of 8.98 \textmu m, and the third place belongs to the random forest with an accuracy of 0.952 and an MAE of 11.41 \textmu m.

We also investigated the accuracy ($R^2$) and MAE results of the ML models in clad depth prediction (Fig. \ref{fig:r2-mae} e-f). Just like the width and height regression results, the highest accuracy is produced by the gradient boosting model which yielded the accuracy of 0.981 and an MAE of 19.96\textmu m in predicting the depth of the clad. The second-best performing ML model for depth prediction was the random forest model with the accuracy of 0.972 and an MAE of 22.43 \textmu m. Moreover, unlike the trend in predicting the other geometrical features, the polynomial regression, with slightly better performance compared to the neural network, was the third best performing model for clad depth prediction with an accuracy of 0.964 and an MAE of 23.41 \textmu m.

\begin{figure*}[ht!]
    \centering
    \includegraphics[width=\linewidth]{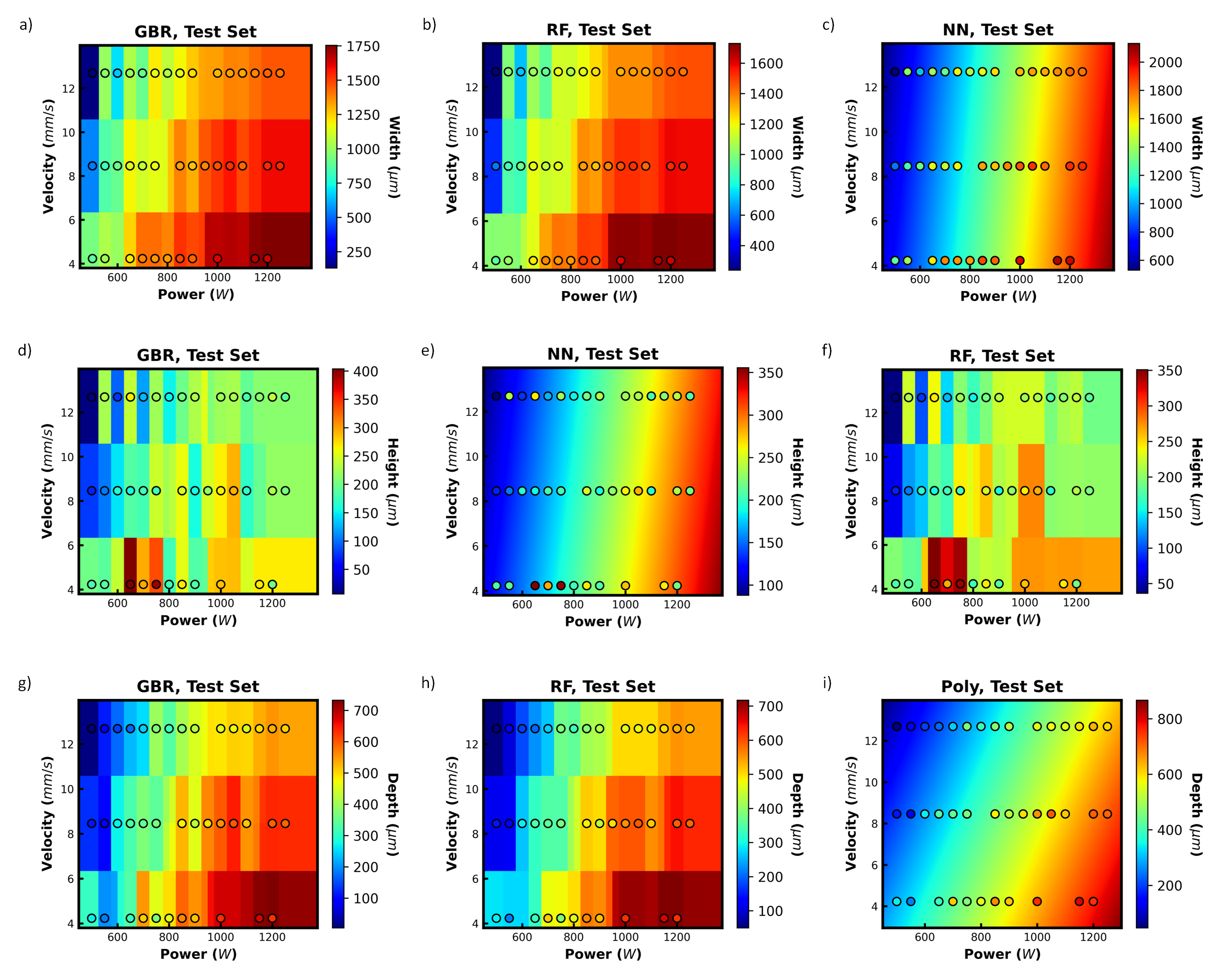}
    \caption{Printability maps constructed by the prediction of ML regression models applied on testing dataset, showing the effect of laser power and laser scanning velocity, on various single clad geometry features: (a,b,c) predicted width using 'Gradient Boosting Regression', 'Random Forest', and 'Neural Network' respectively, (d,e,f) predicted height using 'Gradient Boosting Regression', 'Neural Network', and 'Random Forest' respectively, (g,h,i) predicted depth using 'Gradient Boosting Regression', 'Random Forest', and 'Polynomial Regression' respectively.}
    \label{fig:contour}
\end{figure*}

Figure \ref{fig:r2} presents a comparison between the predicted and ground truth values of the clad geometry during the testing phase, showcasing the performance of the best-performing model for each geometry attribute. A model that exhibits an $R^2$ value of 1 is considered a perfect fit, where all data points lie on the diagonal line $y=x$. The $R^2$ fits displayed in Figure \ref{fig:r2} demonstrate the suitability of the gradient boosting regression models, indicating promising accuracy in predicting the actual geometries. Figure \ref{fig:contour} displays contours of predicted geometry in relation to machine setting features, namely laser power ($w$) and laser scanning velocity ($mm/s$). The points inside the contour represent the test set machine features while the colored areas within the contour range from cold to warm, indicating various predicted geometry values associated with the machine features.

\begin{figure*}[ht!]
    \centering
    \includegraphics[width=\linewidth]{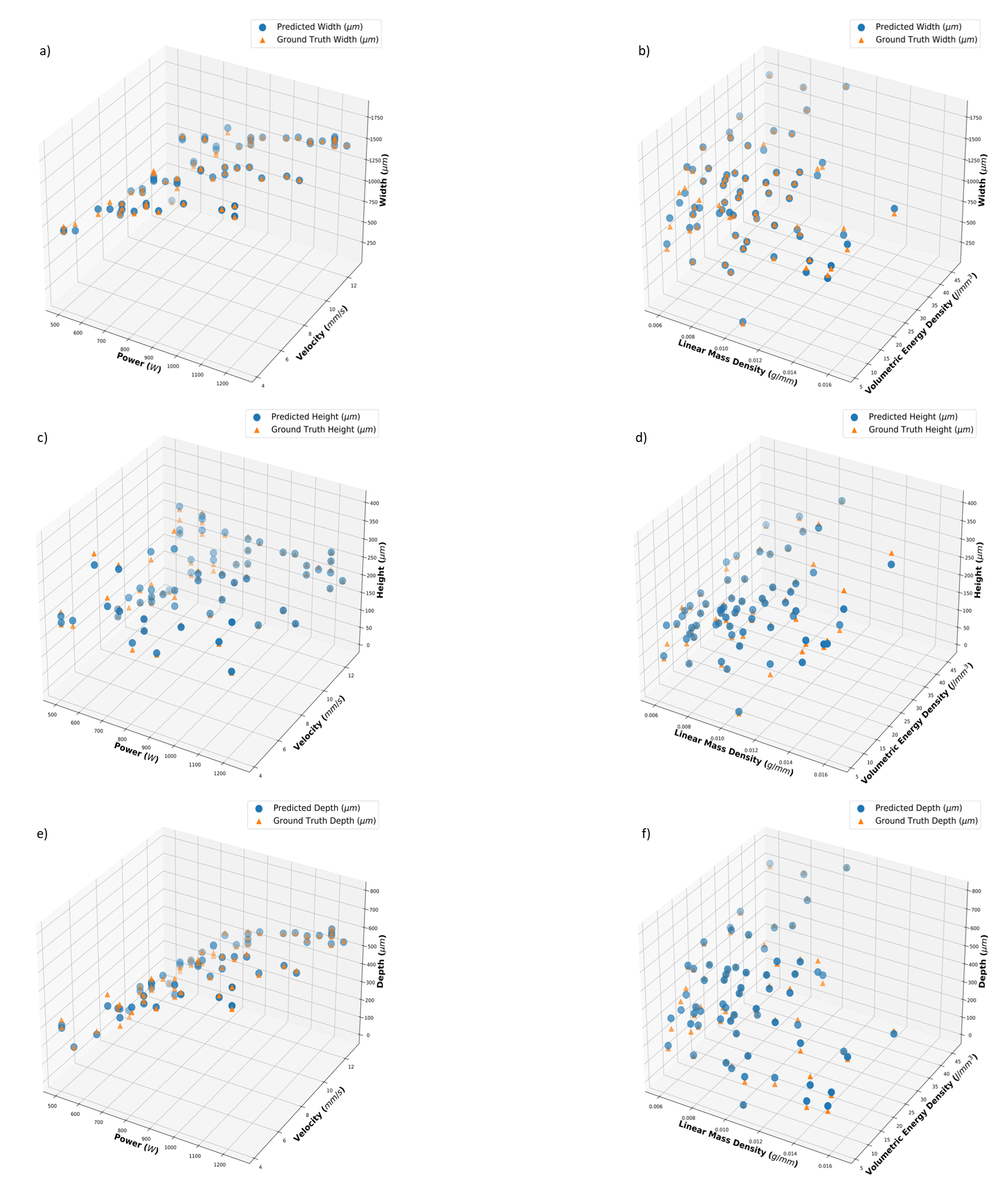}
    \caption{Scatter plots demonstrating the predicted width (a, b), predicted height (c, d), predicted depth (e, f). Predicted and actual data points are plotted with respect to the input processing parameters i.e., Power $(W)$, Velocity $(mm/s)$ and the physics informed features i.e., volumetric energy density $(J/mm^3 )$ and linear mass density $(g/mm)$.}
    \label{fig:pred}
\end{figure*}

After training various ML models for predicting the clad geometrical features i.e., width, height, and depth, the best performing model for each geometry was assessed in the testing phase to compare the values of the predicted geometry with the actual geometry using an unseen subset of the dataset. This comparison is illustrated in Fig. \ref{fig:pred} using three-dimensional scatter plots presenting the predicted clad geometries with respect to the input processing parameters i.e., Power $(W)$, Velocity $(mm/s)$ and the physics informed features i.e., volumetric energy density $(J/mm^3 )$ and linear mass density $(g/mm)$.

\subsection{Process map prediction}

In metal additive manufacturing, the quality of the clad is heavily dependent on machine parameters and the underlying physics of the process, such as the level of porosity and distribution of cellular substructures. Therefore, it is essential to understand how these parameters affect clad quality and whether they produce desired outcomes. Machine learning models provide a suitable solution to address these issues, as they enable the development of data-driven decision-making tools to predict clad quality and guide users in selecting appropriate processing parameters for ideal builds.

For successful training of machine learning classifiers in predicting clad quality and creating an optimal process map, access to preliminary labeled data is critical. In this study, the dilution, one of the measurable features that can describe the quality of a single track for printing fully dense builds, was used to label the required dataset for our ML framework \cite{vaughan2023development}. This approach allowed the creation of a comprehensive dataset that accurately captures the influence of dilution and processing parameters on clad quality. The resulting dataset can then be used to develop and train ML models that accurately predict clad quality under specific processing conditions, thereby facilitating the creation of an optimal process map. By providing a robust, labeled dataset, this approach enhances the accuracy and reliability of the ML models, allowing for more precise prediction of clad quality and ultimately improving the efficiency and cost-effectiveness of the AM process.

\begin{figure*}[ht!]
    \centering
    \includegraphics[width=\linewidth]{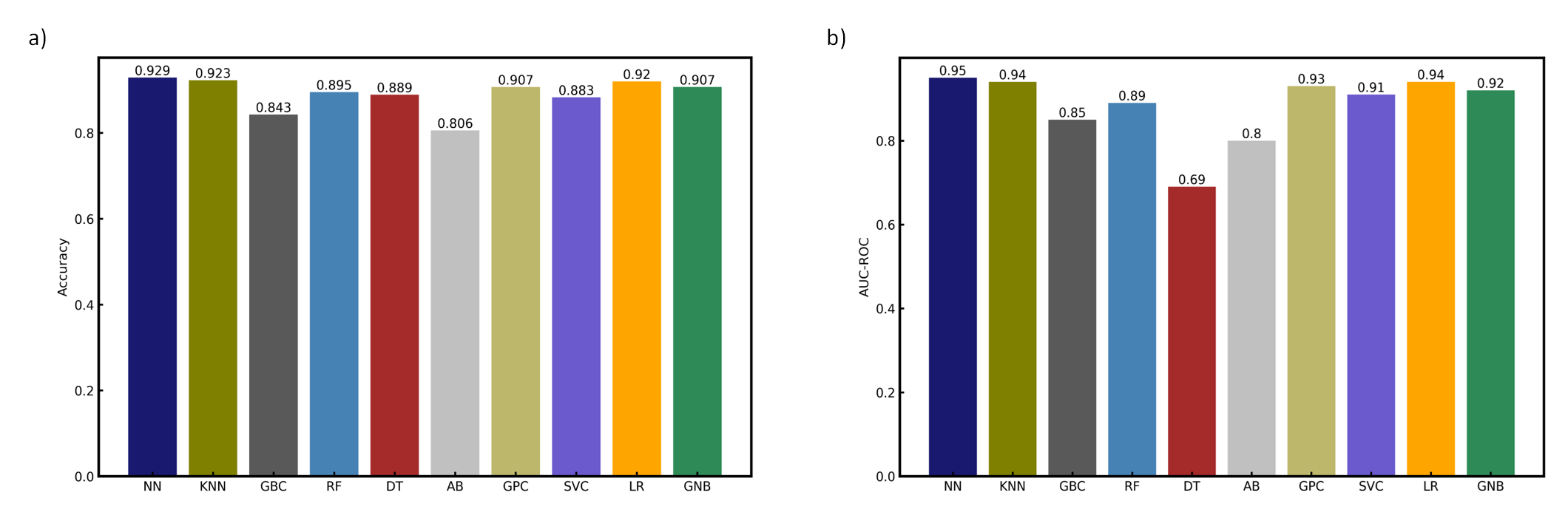}
    \caption{Benchmark performance comparison for predicting the class of the clad and process map: (a) accuracy results, (b) AUC-ROC results.}
    \label{fig:accuracy-auc}
\end{figure*}

In our study, dilution is defined as \cite{erfanmanesh2017empirical, bhardwaj2019direct}:
\begin{equation}\label{eq:17}
Dilution = \frac{Depth}{(Depth+Height)}
\end{equation}
Dilution, a parameter ranging in value from 0 to 1, is a critical indicator of the quality and efficiency of the additive manufacturing (AM) process. A low range of dilution (e.g., 0.10) to zero implies minimal or complete lack of fusion between the melted powder and the substrate. This lack of fusion results in poor quality and structural integrity, as there is no inter-layer bonding when printing multiple layers. Conversely, a high value of dilution (e.g., Dilution $\geq$ 0.5) indicates excessive heat input applied to the melt pool, leading to defects such as keyhole porosity. At the maximum value of dilution (Dilution = 1), only a melt pool is produced by melting the substrate, without any clad being printed. This condition indicates powder evaporation and mass flow problems, which can adversely affect the quality and efficiency of the AM process \cite{bhardwaj2019direct, reichardt2021advances, vaughan2023development}. Additionally, wetting angle $\alpha$ of a single track as an important parameter in describing the quality of a clad is correlated with dilution, where better dilution and smoother track surfaces occur at $\alpha$'s typically greater than 130\degree . Overall, the optimal dilution is selected in the range between 20\% and 50\%, for printing a clad with highest density and minimum defects such as porosity \cite{vaughan2023development}. Therefore, incorporating the criteria of optimal dilution as a parameter in our ML framework enabled us to define the clad quality, and label data points in the training dataset as desirable (20\% $\leq$ $Dilution$ $\leq$ 50\%) or undesirable ($Dilution$ $\leq$ 20\% or $Dilution$ $\geq$ 50\%). This labelling facilitates the training process of precise and efficient prediction models for clad quality and optimal process maps, leading to the creation of high-quality AM parts with improved structural integrity and mechanical properties. By incorporating dilution as a critical parameter in our ML framework, we can train models to predict the quality of the clad based on specific processing parameters, guiding users towards selecting the ideal build conditions.

\begin{figure}[ht!]
    \centering
    \includegraphics[width=\linewidth]{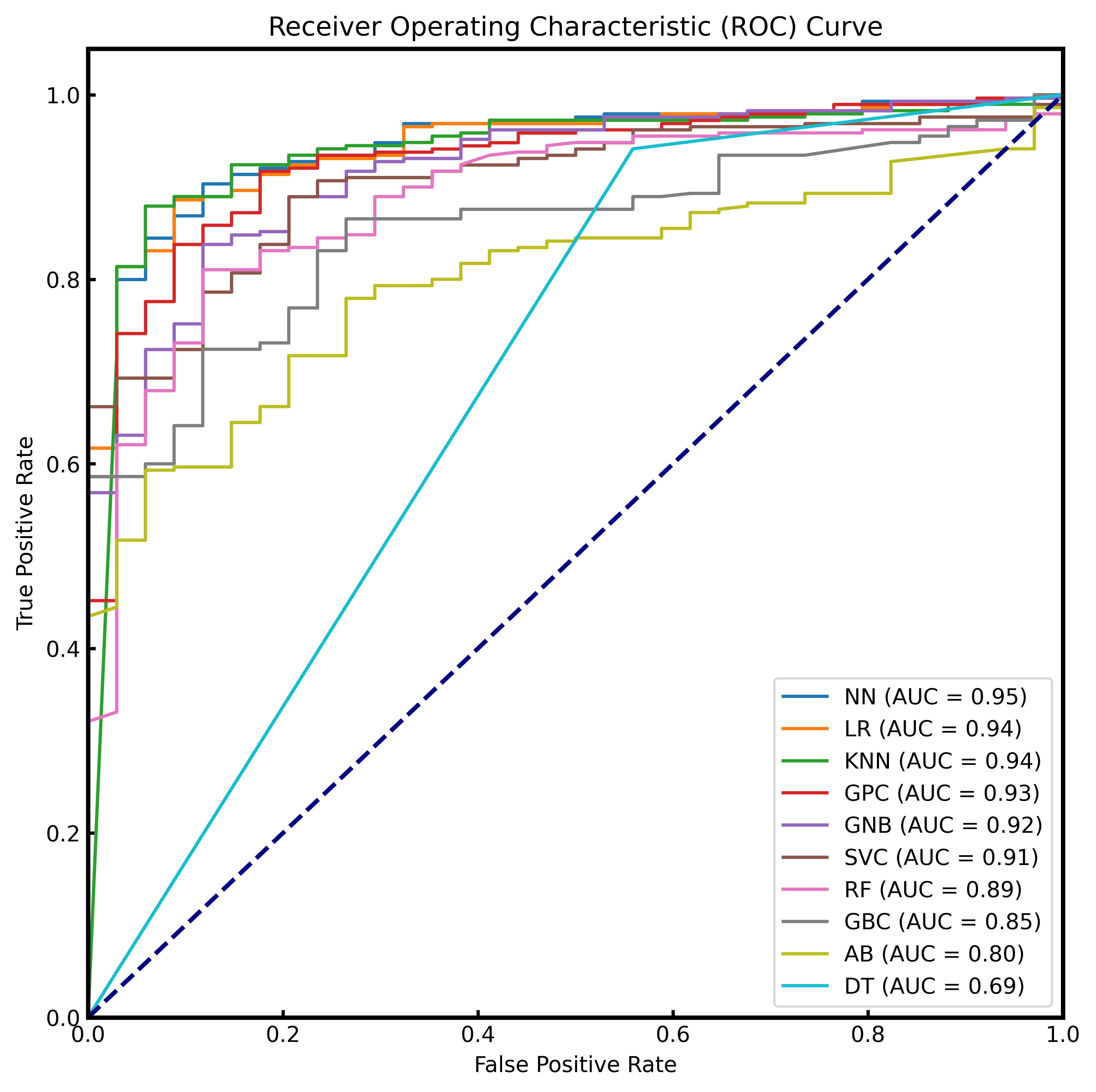}
    \caption{ROC curves of the ML classifiers in predicting the class of the clad and process map.}
    \label{fig:roc}
\end{figure}

Ten different ML models were developed, trained, and tested to classify a clad into desirable or undesirable categories. These models include the ‘Neural Network Classifier’, ‘K-Nearest Neighbors Classifier’, ‘Gaussian Process Classifier’, ‘Logistic Regression’, ‘Gaussian Naïve Bayes’, ‘Support Vector Classifier’, ‘Gradient Boosting Classifier’, ‘AdaBoost Classifier’, ‘Random Forest Classifier’, and ‘Decision Tree Classifier’. The input features for each model were selected from the four features listed in Table \ref{tab:features}, which included machine features and physics-aware features. Hyperparameter optimization was performed to ensure the optimal model with the best performance. To evaluate the performance of the models, accuracy and Receiver Operating Characteristic (ROC) metric were used. ROC curves were plotted to compare the false positive rate and true positive rate of the models, and the area under the curve (AUC) was calculated to determine the performance of each model. The accuracy and AUC-ROC results of the ML classifiers are depicted in Fig. \ref{fig:accuracy-auc} a-b, and the ROC curves of the tested ML classifiers are presented in Fig. \ref{fig:roc}.

\begin{figure*}[ht!]
    \centering
    \includegraphics[width=\linewidth]{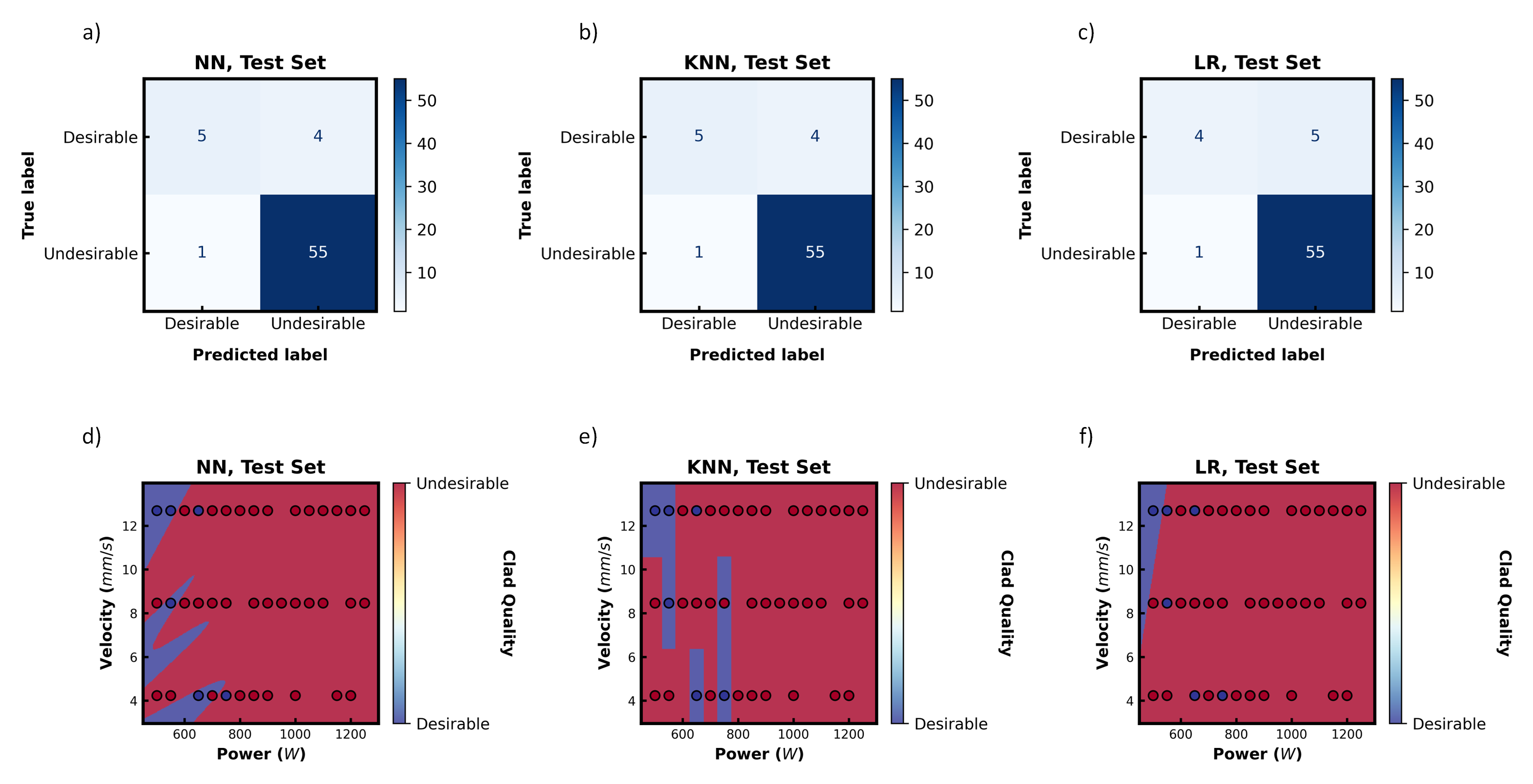}
    \caption{Confusion matrix for clad classification (a) for ‘neural network’, (b) for ‘k-nearest neighbors’, and (c) for ‘logistic regression’. Printability maps constructed by the prediction of ML classification models applied on testing dataset, showing the effect of laser power and laser scanning velocity, on printing a single clad with desirable (20\% $\leq$ $Dilution$ $\leq$ 50\%) or undesirable ($Dilution$ $\leq$ 20\% or $Dilution$ $\geq$ 50\%) quality: predicted process maps using (d) ‘neural network’, (e) ‘k-nearest neighbors’, and (f) ‘logistic regression’}
    \label{fig:confmat-procmap}
\end{figure*}

In this study, we evaluated the performance of ten classifiers for the task of classifying clad as either desirable or undesirable. Our results showed that the neural network, k-nearest neighbors, and logistic regression classifiers were the top-performing models, achieving accuracies of 0.929, 0.923, and 0.920, respectively. Among them, the neural network classifier demonstrated the highest performance in both accuracy and AUC, achieving an AUC of 0.95, while the k-nearest neighbors and logistic regression classifiers followed closely with an AUC of 0.94. These three classifiers demonstrated reliable and robust predictions, highlighting their potential for practical applications in the field of additive manufacturing. The findings of this study provide valuable insights for selecting appropriate classifiers for predicting the quality of clad.

Additionally,we plot the confusion matrices of the three selected best classification models to visualize the performance of these models in labeling the observations. Each row of the confusion matrix represents the occurrence of an observation in the ground truth, while each column represents the occurrence of a class based on the model estimation. The confusion matrices of the selected best classifiers i.e., neural network classifier, k-nearest neighbors classifier, and logistic regression, are depicted in Fig. \ref{fig:confmat-procmap} a - c. Note that the class-0 represents the desired clad, and the class-1 represents the unwanted clad. In addition, visualizing the decision boundaries of the classifiers allows us to assess their behavior in partitioning the testing dataset and predicting the process map that could further be utilized as a powerful tool for data driven decision making in practice. The estimated decision boundaries of the neural network classifier, k-nearest neighbors classifier, and the logistic regression, applied on the testing subset of the dataset are indicated in Fig. \ref{fig:confmat-procmap} d – f with respect to the power and velocity as the independent variables. Blue areas are predicted as the desired process windows that has a high possibility of yielding a desired clad shape, and the red areas are predicted as the unwanted process windows.

\subsection{Model suitability}

The models best suited for the regression task of estimating clad geometry characteristics were gradient boosting, random forest, neural network, and polynomial regression models. Among these models, the gradient boosting model yielded the highest accuracy, outperforming all others. However, both random forest and neural network models showed decent performance with highly reliable predictions despite slightly lower accuracy compared to the gradient boosting model. The polynomial regression model also demonstrated great performance in predicting the depth after the ensemble models and the neural network. These models are highly scalable and robust, making them suitable for expanding and constantly changing data.

In addition to accuracy, it is essential to consider other metrics when selecting models for different applications. The time complexity of each ML model is an important consideration, especially when multiple ML models are utilized in comprehensive frameworks for big-data analysis and predictions. The Big O notation is a widely used mathematical tool for describing the time complexity of algorithms, which is crucial in evaluating the efficiency of different algorithms in handling large data sets. Time complexity pertains to the time it takes for an algorithm to execute as a function of the input data size. The Big O notation \cite{chivers2015introduction} represents the upper bound on the algorithm's time complexity growth rate based on the input size and is useful for comparing the performance of different algorithms, particularly in large-scale data analysis and predictions. Among the chosen models, gradient boosting and random forest, the ensemble models, demonstrated the lowest time complexity, followed by the polynomial regression and neural network models in terms of complexity, respectively. The following section provides the big-O time complexity definitions for each of these models.

\setlength{\tabcolsep}{10pt} % Default value: 6pt
\renewcommand{\arraystretch}{1.2} % Default value: 1
\begin{table*}[ht!]
\centering
\caption{Summary of the three best machine learning models performances and operations time complexity for two tasks i.e., regression task (predicting the clad width, height, and depth), and classification task (predicting the quality label of a clad). (Best performance value is reported in two different metrics for each model, and the time complexity for training each model is illustrated with Big-O notation).}
\label{tab:summary}
\resizebox{12cm}{!} 
{
\begin{tabular}{cccc}
\toprule
Category  &  Best performance  &  Metric  &  Value  \\
\midrule
Width of clad   &  GBR   &   $R^2$ &  0.985 \\
                &        &   MAE   &  24.41 \\
                &        &   Big-O &  $O(knLog(n))$ \\
                &  RF    &   $R^2$ &  0.981 \\
                &        &   MAE   &  32.28 \\
                &        &   Big-O &  $O(knLog(n))$ \\
                &  NN    &   $R^2$ &  0.975 \\
                &        &   MAE   &  29.85 \\
                &        &   Big-O &  $O(mA^2B^2n^2f)$ \\
\midrule
Height of clad  &  GBR   &   $R^2$ &  0.964 \\
                &        &   MAE   &  8.73 \\
                &        &   Big-O &  $O(knLog(n))$ \\
                &  NN    &   $R^2$ &  0.955 \\
                &        &   MAE   &  8.98 \\
                &        &   Big-O &  $O(mA^2B^2n^2f)$ \\
                &  RF    &   $R^2$ &  0.952 \\
                &        &   MAE   &  11.41 \\
                &        &   Big-O &  $O(knLog(n))$ \\
\midrule
Depth of clad  &  GBR    &   $R^2$ &  0.981 \\
                &        &   MAE   &  19.96 \\
                &        &   Big-O &  $O(knLog(n))$ \\
                &  RF    &   $R^2$ &  0.955 \\
                &        &   MAE   &  8.98 \\
                &        &   Big-O &  $O(knLog(n))$ \\
                &  Poly  &   $R^2$ &  0.964 \\
                &        &   MAE   &  23.41 \\
                &        &   Big-O &  $O(n^3)$ \\
\midrule
Clad quality classification  &  NN    &   Classification accuracy &  92.90 \\
                             &        &   AUC-ROC   &  94.59 \\
                             &        &   Big-O &   $O(mA^2B^2n^2f)$ \\
                             &  KNN   &   Classification accuracy &  92.28 \\
                             &        &   AUC-ROC   &  93.78 \\
                             &        &   Big-O &   $O(knd)$ \\
                             &  LR    &   Classification accuracy &  91.97 \\
                             &        &   AUC-ROC   &  94.48 \\
                             &        &   Big-O &  $O(nd)$ \\

\bottomrule
\end{tabular}
}
\end{table*}

\begin{itemize}
\item Gradient boosting and random forest:
\begin{equation}\label{eq:}
O(knLog(n))
\end{equation}
\end{itemize}
where $k$ is the number of trees (estimators), $n$ is the number of training samples, and $d$ is the dimension of the data.

\begin{itemize}
\item Polynomial regression:
\begin{equation}\label{eq:}
O(n^3)
\end{equation}
\end{itemize}
where $n$ is the number of training samples.

\begin{itemize}
\item Neural network:
\begin{equation}\label{eq:}
O(mA^2B^2n^2f)
\end{equation}
\end{itemize}
where $m$ is the number of iterations, $A$ and $B$ are the number of neurons in each of the two hidden layers of our model, $n$ is the number of training samples, and $f$ is the number of features.

The models best suited for the classification task of predicting the desirability or undesirability label of the clad were neural network, k-nearest neighbor, and logistic regression models, which demonstrated great performance in estimating complex decision boundaries in binary classification problems. Among these models, the neural network and k-nearest neighbor models are robust and scalable in an environment where the dataset may constantly expand and become more complex. Additionally, the logistic regression model showed decent accuracy metrics for the classification task, even though it is a fairly simple linear model which estimates a linear decision boundary for classifying the datapoints. Among the selected models, logistic regression has the lowest training time complexity, followed by k-nearest neighbor and neural network. The following shows the big-O time complexity definitions of these models.

\begin{itemize}
\item K-nearest neighbor:
\begin{equation}\label{eq:}
O(knd)
\end{equation}
\end{itemize}
where $k$ is the number of neighbors, $n$ is the number of training samples, and $d$ is the dimension of the data.

\begin{itemize}
\item Logistic regression:
\begin{equation}\label{eq:}
O(nd)
\end{equation}
\end{itemize}
where $n$ is the number of training samples, and $d$ is the dimension of the data. Moreover, the complexity of the neural network model for classification is defined in the same way as for the regression model.

Although Akbari et. al \cite{akbari2022meltpoolnet} and Crammer et. al \cite{crammer2001algorithmic} suggested the suitability of kernel based models, such as SVMs, for the binary classification tasks, the neural network and few other classifiers outperformed the support vector classifier and showed superior performance and lower training time complexity. Moreover, although the Gaussian process methods are suggested for non-linear decision boundaries and demonstrated a decent performance for the classification task in this study, but these models are less feasible in high dimensional environments due to their high computational expense \cite{liu2020gaussian}. A summary of all the selected best performing models for the two explored tasks is provided in \autoref{tab:summary} where the accuracy, error, and time complexity definitions are demonstrated.

In summary, it is crucial to consider not only the accuracy metrics but also the time complexity of the selected ML models when predicting clad characteristics. This is especially important if the framework is intended to be scaled and applied to more complex industrial scale problems in MAM with larger amounts of data.

\section{Conclusion}

In this study, we presented a comprehensive machine learning framework that successfully predicted the clad geometry and optimal process window for additive manufacturing using directed energy deposition. A hybrid method was employed to prepare the dataset by integrating data from both experiments and computational fluid dynamics modeling. Hyperparameter optimization was carried out for each machine learning model, ensuring the highest accuracy in predictions. Our findings revealed that the gradient boosting and random forest models consistently outperformed other models in predicting clad geometrical features, demonstrating superior $R^2$ accuracy and MAE results in the regression task. The neural network model, on the other hand, showed the best performance in predicting the quality label of the single clad, exhibiting the highest accuracy and AUC-ROC results in the classification task.

Furthermore, our study highlights the importance of considering multiple metrics beyond just accuracy when selecting machine learning models for different applications. We emphasize the significance of evaluating the time complexity of each model, especially when using them in large-scale data analysis and predictions. To this end, we employed the Big O notation as a tool to describe the time complexity of each model. Our results showed that the ensemble models, gradient boosting and random forest, had the lowest time complexity, making them highly suitable for handling large data sets. The polynomial regression and neural network models followed in terms of complexity, respectively. Overall, the framework developed in this study provides a practical and effective approach for predicting clad geometry and optimizing the additive manufacturing process using directed energy deposition. We hope that our novel approach and the developed comprehensive ML framework will serve as a foundation for future research in this area and our findings will contribute to the advancement of additive manufacturing technologies. Our study has shown that a hybrid method integrating experimental and simulated data can provide reliable data for machine learning models, which is especially important when the volume of experimental data is limited, and can lead to significant improvements in predictive performance. The next steps in this line of research could include expanding the dataset to include more process parameters, optimizing other ML models, and investigating the transferability of the models to different manufacturing processes.

%%% CRediT authorship contribution statement
\section*{CRediT authorship contribution statement}
\textbf{Sina Tayebati:} Conceptualization, Methodology, Software, Validation, - Formal analysis, Investigation, Data Curation, Writing - Original Draft, Visualization.
\textbf{Kyu Taek Cho:} Conceptualization, Resources, Data Curation, Writing - Review \& Editing, Supervision, Project administration.
%%%

%%%  Declaration of competing interest
\section*{Declaration of competing interest}
The authors declare that they have no known competing financial interests or personal relationships that could have appeared to influence the work reported in this paper.
%%%

%%% Data availability
\section*{Data availability}
The code for this machine learning framework can be accessed at the following link. Additionally, some or all data presented in this study are available from the corresponding author on reasonable request. \\
\url{https://github.com/sinatayebati/CladNet-ML-for-AM}
%%%

%%% Acknowledgments
\section*{Acknowledgments}
The authors gratefully acknowledge NIU for supporting Mr. Tayebati’s study and research in a graduate program at NIU.
%%%

%%% Declaration of AI 
\section*{Declaration of Generative AI and AI-assisted technologies in the writing process}
During the preparation of this work the author(s) used ChatGPT in order to improve the readability and language. After using this tool/service, the author(s) reviewed and edited the content as needed and take(s) full responsibility for the content of the publication.

\bibliography{main}

\end{document}